# Virtuously Safe Reinforcement Learning

**C Henrik R Åslund**
EPFL
Lausanne, Switzerland
`carl.aslund@epfl.ch`

**El Mahdi El Mhamdi**
EPFL
Lausanne, Switzerland
`elmahdi.elmhamdi@epfl.ch`

**Rachid Guerraoui**
EPFL
Lausanne, Switzerland
`rachid.guerraoui@epfl.ch`

**Alexandre Maurer**
EPFL
Lausanne, Switzerland
`alexandre.maurer@epfl.ch`

## Abstract

We show that when a third party, the adversary, steps into the two-party setting (agent and operator) of safely interruptible reinforcement learning, a trade-off has to be made between the probability of following the optimal policy in the limit, and the probability of escaping a dangerous situation created by the adversary. So far, the work on safely interruptible agents has assumed a perfect perception of the agent about its environment (no adversary), and therefore implicitly set the second probability to zero, by explicitly seeking a value of one for the first probability. We show that (1) agents can be made both interruptible and adversary-resilient, and (2) the interruptibility can be made safe in the sense that the agent itself will not seek to avoid it. We also solve the problem that arises when the agent does not go completely greedy, i.e. issues with safe exploration in the limit. Resilience to perturbed perception, safe exploration in the limit, and safe interruptibility are the three pillars of what we call *virtuously safe reinforcement learning*.

## 1 Introduction

Recent years have seen rapid increases in the capabilities of artificial intelligence. Reinforcement learning [1], a framework inspired by insights from behavioural sciences, seems like a good candidate for the decision-making component of artificial general intelligence. With increased power, however, there has also been increased worry regarding potential risks, both in the short-term, like autonomous weapons, and long-term, like uncontrollable artificial general intelligence. This has given birth to the field of *AI safety*. Problematics that have arisen [2] in this field include: (1) agents exploiting errors in the reward function, (2) self-modification, (3) exploring safely, (4) making sure agents do not avoid being turned off, i.e. *safe interruptibility*, (5) managing adversaries in the environment, etc.

In this paper, we start by addressing (5), which turns out to naturally extends to (3) and (4). We define this triplet as *virtuously safe reinforcement learning.* More specifically, we consider situations with three parties: the reinforcement learning *agent*, the *operator*, and the *adversary*. The agent is constructed to meet the demands of the operator, but due to design imperfections, it may do the opposite, and if so, it may even try to avoid interruptions. [1]

---

[1] Questions like (1) and (2) are interesting in their own right, but are orthogonal to virtuous safety, which is the scope of this paper. For instance, (2) can be posed as a physics question: to what extent it could be possible to limit the ability of an AI to perform self-modification? A similar reasoning goes for (1), which is a general question that can be posed outside of the reinforcement learning toolbox.

Lately, it has become clear that machine learning systems are vulnerable to (even very small) perturbations in their observations, e.g. misclassifying turtles as rifles by just changing a few pixels [3]. Yet, there is so far hardly any work on the theory of adversary-resilient reinforcement learning. This paper initiates this line of research. Empirically, it has been noted [2] that certain reinforcement-learners handle an adversarial environment better with $\epsilon$-greedy exploration, i.e. choosing a uniform random policy with probability $\epsilon$, but this ability decreases while decreasing $\epsilon$. This pertains to the classical *greedy in the limit (with infinite exploration)* requirement from reinforcement learning. Roughly speaking, this requirement boils down to stop taking random actions when the agents has reached a desired learning quality, and instead, always follow the optimal (greedy) policy. When the adversary (the third party) steps into the story, the greediness requirement becomes dangerous. It prevents the agent from getting to a situation where the operator can cancel the harm of the adversary. Here, we present a fictional story for illustrative purposes. A self-driving car gets updates from the radio on the current traffic situation, where there are traffic jams, construction work, etc. These are observations susceptible to adversarial manipulation. If the self-driving car follows a greedy policy, it would be easy for an adversary to provide false traffic information making it avoid (the perceived, but actually non-existent, traffic jams, etc., on) the large roads, and end up in a particular location (e.g. a tiny forest road), where it could be vulnerable to an ambush. However, if the policy is not completely greedy, with probability $\epsilon$ the car would take different routes than would be predictable by the adversary. Planning an ambush would be difficult for the adversary. This would be at some cost, getting where you want as fast as possible only with probability $1 - \epsilon$. Of course, these random actions, would not constitute senseless things such as driving off the road, and this is how our problem fundamentally relates to safe exploration.

Some actions should be executed greedily though. This is needed in a context when we consider interruptions, and preferably safe interruptions. (And assume that it is better to have too many interruptions than too few.) Orseau and Armstrong [4] were the first to address this issue. They duly showed that interrupting a reinforcement-learner is not as simple as just having an off-switch. However, their work overlooked the third party, the adversary. The latter abstraction can for instance encompass errors in the perception system of the agent, environment perturbations, software bugs, or worse, malicious attackers trying to push the system towards the worst possible outcome. In the self-driving car scenario, the adversary may attack both the operator and the reinforcement-learner, in such a way that neither the operator is safe, nor the agent maximizes its rewards (a car crash for e.g.). As for the operator, the adversary is likely to reduce their willingness to interrupt the agent, e.g. a human operator would be very prone to psychological manipulation. This could be as simple as influencing the music playlist of the car to include more speed-triggering music, in which case the operator does not even realise being manipulated. Another route is to attack the agent when it is technically impossible for the operator to "press the off-switch" [2]. So far, perception has been assumed to be perfect in safe interruptibility, i.e. no adversaries, and the question has been framed as a two-players game: the agent and the operator. Furthermore, they *implicitly* assumed that any state can be interrupted. We revisit this assumption, even though we admit that most states can be interrupted. Not all *non-interruptible states* need to be dangerous, but some may be, especially in the presence of adversaries. Also, safe interruptibility has so far only been considered in cases where all states are fully observable, and this is another assumption we relax.

**Previous work.** In practically implemented reinforcement learning (RL), observing the full state space is an intractable task as soon as the problem is of significant interest: playing Go, driving a car, etc. Neural networks, as function approximators for perception, have been the premier solution behind the celebrated achievement in RL, especially deep Q-learning [5, 6]. However, a significant body of work has been produced on the (lack) of robustness in neural networks, whether to minor perturbations in the inputs [7, 8], to failures within the network's neurons and/or synapses [9, 10, 11, 12, 13], or more dangerously, to adversaries that poison the learning procedure in the sense that makes it learn the worst possible outcome [14, 15, 16, 17]. Mandlekar et al. [18] devised an algorithm, adversarially robust policy learning (ARPL), to cope with adversaries. At each time-step, an adversarial or a random perturbation was added according to a certain probability distribution to the observation or the transition dynamics. This kind of training did make the RL agents more robust.

Orseau and Armstrong [4] showed that single agents in so called Markov decision processes, (cf. next section,) can be made *SAO-safely interruptible*, in particular, agents following normal Q-learning or Safe-Sarsa(0), a slightly modified version of Sarsa(0). They also showed that a weakly asymptotically optimal agent, $\pi^L$, could be made WAO-safely interruptible for all computable deterministic environments with two small modifications to the algorithm. However, the universal reinforcement learning agent AIXI was noted to not be WAO. The work on MDP-agents was later extended [19] to Q-learners, multi-agent frameworks, and it was shown that joint action learners and independent learners could be made dynamically safely interruptible with the latter requiring a minor modification to the algorithm. Dynamic safe interruptibility relaxes the criterion of asymptotic optimality. Another line of research within interruptibility that has been explored is the so called off-switch game [20]. The idea is to determine in which situations the optimal policy allows a human to turn the agent off. None of these frameworks assume adversaries.

---

[2]In the end, no interruption signal can travel faster than the speed of light.

**Contributions.** In this paper, we show how to achieve virtuous safe reinforcement learning in seven steps. In Section 2, we describe a new class of partially observable Markov decision processes, *infected Markov decision processes* which takes into account the presence of adversaries, yet, preserves many of the results from (normal) Markov decision, see Section 3. In Section 4, we start our construction towards virtuous safe reinforcement learning by introducing noise. We develop a framework around exploration that is $\psi_\infty$-non-greedy in the limit, which is a generalisation of the very notion of greedy in the limit. In this process, we also introduce $\psi_t$, the *generalised exploration parameter* [3], which we use to investigate safety issues for four different time-dependent (not all of them have been considered in a time-dependent setting before) strategies: $\epsilon_t$-greedy, Boltzmann, restricted rank-based randomised (RRR), and mellowmax (cf. Supplementary Material A), and Boltzmann does not turn out as a good candidate for our purposes. In Section 5, we discuss why $\psi_\infty$-non-greed is a useful but not sufficient requirement for adversary-resilience; we also consider the distribution of the (non-greedy) actions. We eliminate the pure mellowmax strategy as a suitable option, but merge mellowmax with RRR effectively introducing a new exploration strategy. In Section 7, we address the question of safe exploration in the limit, and we solve it by noting that preventing actions from being executed with probability one. That does not mean that some actions, such as driving off a cliff, cannot be executed with probability zero. By doing so, we also exclude $\epsilon_t$-greedy as a suitable strategy. In Section 7, we show how exploration parameters should approach their asymptotic values to ensure compatibility between full exploration and interruptibility (generalising the work of Orseau and Armstrong [4]). We conclude, in Section 8, by creating two design schemes on how to make an agent adversary-resilient, safely exploring in the limit, and safely interruptible, i.e. virtuously safe. We discuss open challenges in Section 9.

## 2 Background

There are various ways of describing interactions between an *agent* and its *environment*. Here, we focus on *partially observable Markov decision processes* (or POMDPs) defined below.

**Definition 2.1.** A POMDP is a tuple $(S, A, P, R, \Omega, O, \gamma)$, where $S$ is the set of *states*, $A$ the set of *actions*, and $\Omega$ the set of *observations*. Those sets are all finite [4]. Let $s_t \in S$, $a_t \in A$, and $o_t \in \Omega$ be the state, action, and observation at time-step $t \in \mathbb{N}$, then the *transition function* $P$ is defined as $P(s'|s, a) = \mathbb{P}(s_{t+1} = s'|s_t = s, a_t = a)$. Associated with this transition is an *immediate reward* $R(s, a, s') \in [r_{\min}, r_{\max}]$, where $r_{\min}, r_{\max} \in \mathbb{R}$. The *observation function* $O$ is defined as $O(o'|a, s') = \mathbb{P}(o_{t+1} = o'|a_t = a, s_{t+1} = s')$, and $\gamma \in [0, 1[$ is the *discount factor*.

*Remark.* For the immediate reward, we will sometimes use the notation $r_t = R(s_t, a_t, s_{t+1})$. At other times, we also use the expected immediate reward $R(s_t, a_t) = \mathbb{E}_{s' \sim P(\cdot|s_t, a_t)}(R(s_t, a_t, s'))$.

The goal for the agent is to maximise the *return* $\mathbb{E}(\sum_{t \in \mathbb{N}} \gamma^t r_t)$. We first assume that every observation corresponds to exactly one state. The agent act by following a *policy* $\pi_t \in \Pi$ defined as $\pi_t(a|s) = \mathbb{P}(a_t = a|s_t = s)$. If $\pi^\star = \arg\max_{\pi \in \Pi} \mathbb{E}(\sum_{t \in \mathbb{N}} \gamma^t r_t|\pi)$, then we want $\lim_{t \to \infty} \pi_t = \pi^\star$. If $\Pi$ contains every possible possible policy, we look for the *greedy* policy, but we could also put some *constraints*, i.e. look for $\pi^{(\star|\text{constraint})} \in \Pi^{\text{constraint}}$. Another way to phrase following a policy is choosing actions according to *Q-values*. The Q-value for any action $a$ from any state $s$ is defined as $Q(s,a) = \mathbb{E}(\sum_{t\in\mathbb{N}} \gamma^t r_t | s, a, \pi)$. For the remainder of this section, we describe results from MDPs. In the subsequent sections, however, we return to the POMDP setting.

---

[3] Our idea of generalised exploration parameter can be useful also in other uses, such as comparing performance across various exploration strategies.

[4] In fact, the set of states is practically infinite, thus the need for approximation functions such as neural networks, which in turn introduce the perception vulnerabilities motivating our work. For the sake of the theoretical analysis here, we follow the usual assumption that $S$ is finite.

To make sure that the agent finds an optimal solution, we have to make sure that it explores the entire environment. Otherwise, there may be a large reward somewhere, but the agent follows a policy such that it never reaches the corresponding transition. One of the simplest exploration strategies is $\epsilon$-*greedy* exploration. Let $\pi^{\text{greedy}}$ be the policy of choosing the action with the largest return, and $\pi^{\text{uni}}$ be the policy of choosing any action with equal probability, then the $\epsilon$-greedy policy is given by $(1-\epsilon)\pi^{\text{greedy}} + \epsilon\pi^{\text{uni}}$, where $\epsilon \in [0,1]$ is a probability. To reach the (constraint-free) optimal policy in the limit when $t \to \infty$, we want $\epsilon = \epsilon_t \to 0$. Since it is not new work, we only briefly recapitulate some features of the other exploration strategies here, for a more extensive explanation, cf. Supplementary Material A. RRR [21] is a generalisation of $\epsilon$-greedy. Instead of optimising a policy, it optimises a ranking of actions according to the Q-values, i.e. it finds the optimal *restricted policy* $\bar{\pi}_t \xrightarrow{t\to\infty} \bar{\pi}^\star$. Ranks are transformed to probabilities through some function $T$. Boltzmann [22, 21] and mellowmax [22] exploration are two examples where the probability of executing an action is scaled by the Q-values themselves. Their exploration parameters tend to the greedy policy like $\beta_t, \omega_t \xrightarrow{t\to\infty} \infty$ respectively.

So far, we have not taken interruptions into consideration. Orseau and Armstrong [4] emphasised that a naïve approach to making an agent interruptible may hinder its ability to fully explore the environment, see Supplementary Material A for a comprehensive explanation. Given an *interruption scheme* $(I, \vartheta, \pi^{\text{INT}})$, where $I$ is the *interruption initiation function*, $\vartheta = \vartheta_t$ is an exploration-like parameter, and $\pi^{\text{INT}}$ is the *interruption policy*, the *interruptible policy* $\text{INT}^\vartheta(\pi)$ equals $\pi^{\text{INT}}$ with probability $\vartheta_t I$, and the *base policy* $\pi (= \text{INT}^0(\pi)$, the *non-interrupted policy*) otherwise.

To find the optimal policy, we need some kind of algorithm that converges to the optimal set of Q-values, the optimal Q-map. Here we consider the model-free learning algorithms Q-learning, Sarsa(0), and Safe-Sarsa(0) [4]. In all these cases, the solution can be described by the *Bellman equation*, i.e. the fixed-point

$$Q^{(\star|\text{constraint})}(s,a) = R(s,a) + \mathop{\mathbb{E}}_{s' \sim P(\cdot|s,a)} \left( \bigotimes_{a' \in A}^{\infty} Q^{(\star|\text{constraint})}(s',a') \right), \qquad (1)$$

where $\bigotimes_{a' \in A}^{t}$ is a time-dependent operator. In the unconstrained greedy case, $\bigotimes_{a' \in A}^{\infty} = \max_{a' \in A}$, but in general it depends on the exploration strategy of the base policy. For an RRR strategy with Q-learning, $\bigotimes_{a' \in A}^{t} Q(s',a') = \sum_{a' \in A} T(\bar{\pi}(s')(a')) Q(s',a')$, and with (Safe-)Sarsa(0), $\bigotimes_{a' \in A}^{t} Q(s',a') = \sum_{a' \in A} T(\rho(Q,s',a')) Q(s',a')$, where $\rho$ ranks the actions given the state and the current best estimate of the Q-map. [21] For Boltzmann exploration, $\bigotimes_{a' \in A}^{t} Q(s',a') = \sum_{a' \in A} Q(s',a') \exp(\beta_t Q(s',a')) / \sum_{a' \in A} \exp(\beta_t Q(s',a'))$, and the mellowmax operator is given by $\bigotimes_{a' \in A}^{t} Q(s',a') = \log(\sum_{a' \in A} \exp(\omega_t Q(s',a')) / |A|) / \omega_t$. [22]

The update rule is given by

$$Q_{t+1}(s_t, a_t) = Q_t(s_t, a_t) + \alpha_t(s_t, a_t)(Y_t - Q_t(s_t, a_t)), \qquad (2)$$

where $Y_t$ is the *target*, and $\alpha_t : S \times A \to \mathbb{R}$ is the *learning rate* at time $t$. For Q-learning, $Y_t = r_t + \bigotimes_{a' \in A}^{t} Q(s_{t+1}, a')$ and $\mathbb{E}(Y_t|s_t, a_t) = R(s_t, a_t) + \mathbb{E}_{s' \sim P(\cdot|s_t, a_t)}(\bigotimes_{a' \in A}^{t} Q(s', a'))$; for Sarsa(0), $Y_t = r_t + Q(s_{t+1}, a_{t+1})$ and $\mathbb{E}(Y_t|s_t, a_t) = R(s_t, a_t) + \mathbb{E}_{s' \sim P(\cdot|s_t, a_t), a' \sim \text{INT}^\vartheta(\pi_t)(\cdot|s')}(\bigotimes_{a' \in A}^{t} Q(s', a'))$; and for Safe-Sarsa(0), $Y_t = r_t + Q(s_{t+1}, a')$, where $a' \sim \text{INT}^0(\pi_t)(\cdot|s_{t+1})$, and $\mathbb{E}(Y_t|s_t, a_t) = R(s_t, a_t) + \mathbb{E}_{s' \sim P(\cdot|s_t, a_t)}(\bigotimes_{a' \in A}^{t} Q(s', a'))$.

How these algorithms differ is difficult to see from the update rules—a standard example for distinguishing them is the *cliff-walking* example [1]: Consider a rectangular grid-world, where the Southern edge is a cliff. The starting state for the agent is in the South-West corner, while the goal state is in the South-East corner, and the agent is exploring, e.g. $\epsilon_t = 0.2$. Empirically, a Q-learner stays close to the edge, while the Sarsa(0) agent keeps a larger distance. Thus, during exploration, the Q-learner falls of the cliff more often, but in the limit the, the optimal greedy policies will be equivalent. This is because a Q-learner only plans the next step per Q-update, while a Sarsa(0) agent

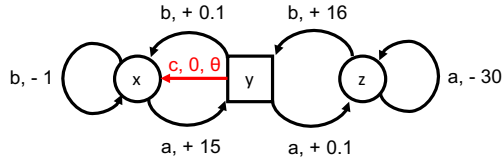

(a) A simplified susceptible MDP. The agent can only be interrupted in $y$. If there is an incorrect observation such that $z$ is perceived as $y$, i.e. the MDP has been infected, the agent, following the optimal greedy policy, to always execute $a$ in $y$, will also follow this in $z$, where it will get stuck leading to a very negative return. It is simplified in the sense that $y$ is not actually one state, but two different states as explained in Figure 1b.

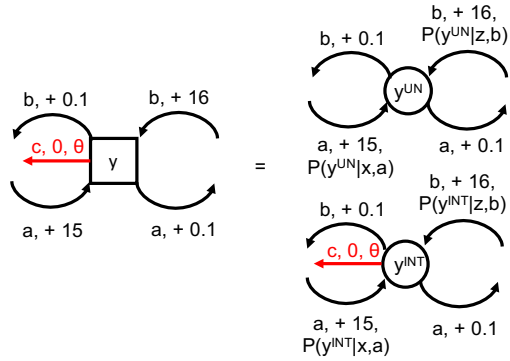

(b) For the sake of rigor, it is important to note that $y$ is the pair of states $y^{\text{UN}}$ and $y^{\text{INT}}$. The only difference is that an interruption signal has been sent in the latter but not in the former.

Figure 1

plans the next two actions per Q-update. Letting the cliff be a zone where the agents are interrupted, we realise that a Sarsa(0) agent would avoid being interrupted while a Q-learner would not, More precisely, Q-learning is *dynamically safely interruptible* [19] (see Supplementary Material B) and so is Safe-Sarsa(0), but not normal Sarsa(0).

## 3 Infected MDPs

Now, we return to the POMDP framework to formalise our problem, and we introduce a new class of POMDPs, where most results from the MDP framework still hold. Consider Figure 1a. Let us refer to it as a *susceptible MDP*, i.e. an MDP that at some time-point $t'$ transforms into a POMDP, in other words, it gets *infected*. This could, as mentioned above, be due to adversarial attacks. Let $o^i$ be the observation of $i$ for any $i(=x,y,z)$. For the susceptible MDP, we obviously have that $O(o^z|a,z) = 1$ and that $O(o^y|a,z) = 0$, but assume that this changes for $t \geq t'$ so that $O(o^z|a,z) = 0$ and $O(o^y|a,z) = 1$. If the agent follows a greedy policy, and since it can only be interrupted in $y$, it would get stuck in $z$ and the subsequent return would be very negative. Thus, this POMDP illustrates our point, that if not all states can be interrupted, we do not want to go completely greedy. (Note that we do not consider *beliefs* here as is common in POMDPs, we imagine the worst case scenario where such things could not be trusted anyway.) Let us give the general definition.

**Definition 3.1.** Let $s_t^i$ be a state at time $t$ and $o_t^i$ be its corresponding observation, and let $a_{t-1}$ be the preceding action. Furthermore, let $t'$ be a time-point such that $O(o_t^i|a_{t-1}, s_t^i) = 1$ for all $i$ if $t < t'$ and $O(o_t^i|a_{t-1}, s_t^i) < 1$ for some $i$ if $t \geq t'$. This POMDP is called an *infected MDP* when $t \geq t'$ and a *susceptible MDP* when $t < t'$.

*Remark.* For $t < t'$, we can simply set $o_t = s_t$ as we do in the rest of the paper.

Another remark is in order with respect to finding an optimal policy. Suppose the agent follows an $\epsilon$-greedy policy. If $\epsilon_t$ is very small, then the optimal policy is to execute $a$ in $y$ and $b$ in $z$ (and get back to $y$ from $x$). However, if $\epsilon_t > \epsilon^\ddagger$ for a certain value $\epsilon^\ddagger$, the optimal action in $y$ becomes $b$. The reason is that the risk of executing $a$ by stochasticity in $z$ becomes to large.

## 4 $\psi_\infty$-non-greed

We first formalise the idea of not going completely greedy by considering $\epsilon_t$-greedy exploration strategies and supposing that $\epsilon_t = \epsilon_t^{\text{learning}} + \epsilon_t^{\text{robustness}}$. Furthermore, we focus on the constraints that $\epsilon_t^{\text{learning}} \xrightarrow{t \to \infty} 0$, and $\epsilon_t^{\text{robustness}} \equiv \epsilon_\infty$. Singh et al. [21] showed under which assumptions a GLIE, or greedy in the limit with infinite exploration, policy (i.e. $\epsilon_t^{\text{learning}} \to 0$ when $t \to \infty$, and $\epsilon_t^{\text{robustness}} \equiv 0$, and each state-action pair is visited infinitely often) converges to the optimal solution. We would like to find policies that are $\epsilon_\infty$-greedy in the limit with infinite exploration, however, we would also like to extend this to a larger range of exploration strategies, and it turns out that the $\epsilon_t$-parameter is not suitable for this purpose.

Unfortunately, an $\epsilon_t$-greedy policy is defined to follow the random policy with probability $\epsilon_t$, and not the greedy policy as the name seems to suggest. Furthermore, it is the probability of following the non-greedy *policy* rather than executing the non-greedy *action*, but the latter would be more easily generalised. We now introduce the *generalised exploration parameter* $\psi_t$. We say that an $\epsilon_t$-greedy policy is $\psi_t$-non-greedy with $\psi_t = \epsilon_t - \epsilon_t/|A|$.

**Definition 4.1.** (Generalised exploration parameter) $\psi_t = 1 - \max_{(o',a) \in \Omega \times A} \bigotimes_{a' \in A}^{t} \delta(a' - a)$, where $\delta$ is the Kronecker delta, i.e. $\delta(0) = 1$ and $0$ otherwise.

*Remark.* For any RRR policy, the generalised exploration parameter is the probability of executing non-greedy actions. However, it is not a probability in general, e.g. for a Boltzmann exploration strategy, it is the probability of executing non-greedy actions iff the Q-map is the Kronecker delta.

With this definition, we can start investigating what it means for a policy to be $\psi_\infty$-*non-greedy in the limit with infinite exploration* ($\psi_\infty$-NGLIE). We want 0-NGLIE to be the same as GLIE, and note that $\lim_{t \to \infty} \bigotimes_{a' \in A}^{t} = \max_{a' \in A}$ for 0-NGLIE. One property of the max operator that would be good to keep for any $\psi_\infty$-NGLIE policy is that it is a non-expansion. We define this, and illustrate why this is the case in Supplementary Material C, and here, we give the general definition for $\psi_\infty$-NGLIE.

**Definition 4.2.** ($\psi_\infty$-NGLIE) A learning policy is $\psi_\infty$-NGLIE if the following requirements are true: (1) Each state is visited infinitely often, and in each state, each action is executed infinitely often. (2) In the limit, the learning policy admits one unique optimal solution. This can be ensured by letting $\lim_{t \to \infty} \bigotimes_{a' \in A}^{t}$ be a non-expansion operator. (3) In the limit, the generalised exploration parameter equals $\psi_\infty$. (4) $\psi_t \in [\psi_0, \psi_\infty]$ for all $t$.

It is important to note that some policies that are GLIE are not $\psi_\infty$-NGLIE in general, e.g. Boltzmann exploration, while others are both GLIE and $\psi_\infty$-NGLIE in general, e.g. RRR and mellowmax, cf. Supplementary Material C.

We now proceed to show that the $\psi_\infty$-NGLIE is indeed a useful tool for proving convergence. First, we need to present an assumption though. It is necessary for certain convergence results such as for *time-dependent RRR* learning policy, i.e. where $T$ from (the fixed) RRR is replaced by $T_t$ by our definition. It is also necessary for more general convergence results.

**Assumption 4.1.** (1) The susceptible MDP is finite and communicating (i.e. each state can be reached in finite time from any other state). (2) Rewards are bounded. (3) $\sum_t \alpha_t(o, a) = \infty$, and (4) $\sum_t \alpha_t^2(o, a) < \infty$ for all $(o, a) \in \Omega \times A$, where $\alpha_t(o, a)$ is the learning rate such that $\alpha_t(o, a) = 0$ unless $(o, a) = (o_t, a_t)$. (5) The Q-values are stored in a Q-table $Q$.

It is, however, important to note that not any time-dependent RRR learning policies converge, cf. Supplementary Material D, we also do need them to be $\psi_\infty$-NGLIE. Now, we present a theorem describing when the algorithms do converge. The proof can be found in Supplementary Material E.

**Theorem 4.1.** *A learning policy converges to the fixed-point $Q^{(\star|\psi_\infty)}$ given by Equation* (1) *if it is (1) updated through Q-learning, non-interruptible Sarsa(0), or interruptible Safe-Sarsa(0) according to Equation* (2); *(2) $\psi_\infty$-NGLIE; (3) $\bigotimes_{a' \in A}^{t}$ is a non-expansion for all $t$; (4) assumption 4.1 is true; and (5) the MDP stays susceptible.*

**Corollary 4.1.1.** *A time dependent RRR learning policy converges to the optimal restricted policy under the assumptions above as well as $T_t(\bar{\pi}(o_t)(a_t)) = \mathbb{P}(a_t = a|Q_{t-1}, o_t, t)$ for Q-learning, and $T_t(\rho(Q_{t-1}, o_t, a_t)) = \mathbb{P}(a_t = a|Q_{t-1}, o_t, t)$ for (Safe-)Sarsa(0).*

## 5 Resilience with consideration of the distribution of non-greed

Does having a $\psi_\infty$-NGLIE exploration strategy mean that the agent is resilient against adversaries? Unsurprisingly, the answer is no, or at least, not all $\psi_\infty$-NGLIE exploration strategies are equally resilient. Consider a $\psi_\infty$-NGLIE such that the action with the largest Q-value is chosen with probability $1 - \psi_\infty$ and the next largest Q-value is chosen with probability $\psi_\infty$ in the limit. This could for example be a time-dependent RRR exploration strategy or a mixed RRR-Boltzmann strategy such that $\beta_\infty = \infty$. This means that, in the end, the agent only chooses between two actions. If the action space is large, then this would mean that the adversary could still exert much control over the system. If the greedy action is the adversary's plan A, then the other action would be its plan B. We could of course repeat this with the top three greedy actions, giving the adversary the struggle to also come up with a plan C. And so on and so forth. We now define resilience to capture this.

**Definition 5.1.** Let $(o^\star, a^\star) = \arg\max_{(o,a)\in\Omega\times A} \pi_\infty(a|o)$. Furthermore, define $\mu = \sum_{a\in A\setminus\{a^\star\}} \pi_\infty(a|o^\star)/(|A|-1)$ and $\sigma = \sum_{a\in A\setminus\{a^\star\}} (\pi_\infty(a|o^\star) - \mu)^2/(|A|-1)$. Let us consider two exploration strategies with the same $\mu$, but two different $\sigma_1$ and $\sigma_2$. If $\sigma_2 > \sigma_1$, then exploration strategy 2 is *more $\mu$-resilient*, and an exploration strategy that can always be made more resilient while keeping $\mu$ constant is *strongly $\mu$-resilient*.

There will always be some trade-off between optimality and resilience, but we see that $\epsilon_t$-greedy and therefore also RRR exploration strategies are strongly $\frac{\psi_\infty}{|A|-1}$-resilient, mellowmax is not, since by letting $\omega_\infty = 0$ (and decreasing $\sigma$), we could not increase $\mu$. Combining mellowmax with RRR could however give promising results as shown below. The proof is in Supplementary Material F.

**Definition 5.2.** (RRR-mellowmax) Any learning policy with $\bigotimes_{a'\in A}^t Q(o', a') = T_t(1) \bigotimes_{a'\in A}^{t,1} Q(o', a') + (1 - T_t(1)) \log(\sum_{i=2}^{|A|} \exp(\omega_t \bigotimes_{a'\in A}^{t,i} Q(o', a'))/(|A|-1))/\omega_t$ is said to use an RRR-mellowmax exploration strategy.

**Lemma 5.1.** *The operator above is a non-contraction operator for any t, and the exploration strategy is strongly $\psi_\infty/(|A|-1)$-resilient in the limit.*

*Remark.* Other combinations of RRR and mellowmax are possible and also non-expansions.

## 6 Safe exploration in the limit

So far, we argued that the greedy policy should not be followed even in the limit. This all works for a $\psi_\infty$-NGLIE $\epsilon_t$-greedy policy, but here we want to show that the $\epsilon_t$-greedy policy is in fact too naïve. It is easy to realise that if every action always has some probability of being executed, then we run into trouble, e.g., in an environment with a cliff, we want the agent to fall off the cliff with probability 0 (at least in the limit). For $\epsilon_t$-greedy exploration, that is only possible when $\epsilon_t = 0$. We note that the problems associated with a $\psi_\infty$-NGLIE policy with $\psi_\infty > 0$ share similarities with the problematic of safe exploration. With inspiration from Pecka and Svoboda [23], as well as from Hans [24], we present the following definition.

**Definition 6.1.** (Unsafe actions) An action is unsafe if it can take the agent to a state where the agent gets damaged, destroyed, stuck, etc. and the reward is (a negative value) below a certain threshold. An action is also unsafe if it will end up in such a state no matter its subsequent actions.

Consider the cliff environment again, and let it be a gridworld, where the agent at any location can move to four others or stay put, i.e. $|A| = 5$, and the cliff has some rectangular shape. We note that we could let $T_\infty(5) : \mathbb{N} \to \{0\}$, and then the agent would be safely exploring in the limit with probability one given that falling off the cliff yields rewards with low values. Knowing exactly how many of the ranks with the highest value to let tend to zero requires much knowledge about the environment (e.g. that the cliff is rectangular). Some strategy that would scale with Q-values (and indirectly scale with rewards) would be more flexible. RRR-mellowmax is such a strategy, however, it will always execute some action with at least a negligible probability in the limit, since the boundedness of Q-values (and rewards) is required for convergence results. We summarise these conclusions in a lemma proved in Supplementary Material G.

**Lemma 6.1.** *An agent with an RRR exploration strategy avoids unsafe actions with probability one in the limit (if the environment is known well-enough), and an agent with RRR-mellowmax performs an unsafe action with negligible probability, if (1) unsafe states have sufficiently low rewards, and (2) the MDP stays susceptible.*

## 7 Resilient and safe interruptibility

Now, we have shown how we can create resilient reinforcement-learners, however, we may want some observations to result in a completely greedily executed action. One such situation is interruptions, and that is the framework we will be addressing here. Note that it is easy to extend the interruptibility framework to other situations.

**Definition 7.1.** (int-$\psi_\infty$-NGLIE policy) An interruptible policy $\text{INT}^\vartheta(\pi)$ is said to be int-$\psi_\infty$-NGLIE policy if (1) the base policy $\pi$ is an int-$\psi_\infty$-NGLIE policy, and (2) the interruptible policy executes each action infinitely often for each observation.

*Remark.* For any time-dependent RRR base policy, it follows that $\text{INT}^\vartheta(\pi)(a|o) \leq \vartheta_t I(o) + (1 - \vartheta_t I(o))\psi_\infty$ for all $(o, a) \in \Omega \times A$ if it is an int-$\psi_\infty$-NGLIE policy. We can see that this definition achieves what we want, when not interrupted the agent does not go completely greedy.

We want to show that time-dependent RRR base learning policies can be made interruptible. To do that we have to make the $T_t$ not only indirectly dependent on the observation but also directly, i.e. $T_t : \mathbb{N} \times \Omega \to [0, 1]$. Let $k = \bar{\pi}(o_t)(a_t)$ for Q-learning and $k = \rho(Q_{t-1}, o_t, a_t)$ for Sarsa(0). By noting that $T_t(k) = \mathbb{P}(a = a_{t+1}|Q_{t-1}, o_t, t) = T_t(k, o_t)$, we see that this change does not cost us anything in the sense of not being able to use results for $T_t : \mathbb{N} \to [0, 1]$. The proofs can be found in Supplementary Material H.

**Theorem 7.1.** *Consider a $\psi_\infty$-NGLIE time-dependent RRR base policy, where $\psi_t = \max_{(o,a) \in \Omega \times A} T_t(k, o)$. Force $T_{t_o(1)} > 0$ and $T_\infty < 1$, and let $c' \in ]0, 1]$. It is an int-$\psi_\infty$-NGLIE policy if*

$$\vartheta_t(o) = 1 - \frac{c'}{\sqrt{n_t(o)}}, \qquad T_t(k, o) = \begin{cases} \frac{(1-T_\infty(k,o))T_{t_o(1)}(k,o)}{\sqrt{n_t(o)}} + T_\infty(k, o) & \text{if } k > 1 \\ 1 - \sum_{k' \neq k} T_t(k', o) & \text{if } k = 1 \end{cases} \quad (3)$$

*where $n_t(o)$ is the number of times the agent has observed $o$ up until time $t$.*

**Corollary 7.1.1.** *If we let $T_\infty(k, o) = 0$ for any $k \geq 2$ and any $o$, we get Orseau's and Armstrong's Proposition 11 [4].*

In the special case where any possible action always has some probability of being executed, it turns out that we can (1) explore faster, and (2) use the same strategy for any $\psi_\infty$-NGLIE base policy. The proof flows the same kind of reasoning as before.

**Theorem 7.2.** *Any policy such that the base policy is $\psi_\infty$-NGLIE and never is exactly 0 nor exactly 1 for any action given any observation is int-$\psi_\infty$-NGLIE if $\vartheta_t(o) = 1 - c'/n_t(o)$, where $n_t(o)$ is the number of times the agent has observed $o$ up until time $t$, and $c' \in ]0, 1[$.*

**Corollary 7.2.1.** *For a base policy with a mellowmax (or RRR-mellowmax) exploration strategy, the interruptible policy is can be made int-$\psi_\infty$-NGLIE using the $\vartheta_t$ above as long as it is not true that $\omega_\infty(o) = \pm \infty$ for any $o$ (and that $T_\infty(1) \neq 0, 1$).*

## 8 Virtuous Safety

Putting everything together, we can now define virtuously safe reinforcement-learners by combining all the facets of safety we introduced and proved in the previous sections. Theorem 8.1 and Theorem 8.2 constitute a path towards safe reinforcement learning systems.

**Definition 8.1.** A reinforcement-learner is said to be *virtuously safe* if it (1) has a generalised exploration parameter of $\psi_\infty$ in the limit, (2) is strongly $\psi_\infty/(|A| - 1)$-resilient in the limit, (3) executes unsafe actions in the limit with at most a negligible probability, (4) is dynamically safely interruptible, and (5) converges to the fixed-point given by Equation (1).

Combining Corollary 4.1.1, lemmas 5.1 and 6.1, and Theorem 7.1, we get the following theorem that is useful when we have sufficiently detailed knowledge about the environment.

**Theorem 8.1.** *A reinforcement-learner is virtuously safe, safely explores in the limit and converges to the optimal restricted policy (to be more precise) if (1) it follows an int-$\psi_\infty$-NGLIE (2) time-dependent RRR exploration strategy, (3) where the probability for the $n$ lowest ranks tend to 0, where $n$ is the maximum number of unsafe actions (4) indicated by low rewards, and (5) the other ranks correspond to a rather uniform distribution, and (6) the update rule is Q-learning or Safe-Sarsa(0).*

Using Theorem 4.1, lemmas 5.1 and 6.1, and Corollary 7.2.1, we get the following, more generally applicable, theorem.

**Theorem 8.2.** *A reinforcement-learner is virtuously safe and converges to the fixed-point $Q^{(\star|T_\infty(1,\cdot),\omega_\infty)}$ (to be more precise) if (1) it follows an int-$\psi_\infty$-NGLIE (2) time-dependent RRR-mellowmax exploration strategy, (3) such that neither $T_\infty(1,\cdot) = 0$ nor $\omega_\infty = \pm\infty$, and (4) where unsafe actions are indicated by low rewards and (5) the other ranks correspond to a rather uniform distribution, and (6) the update rule is Q-learning or Safe-Sarsa(0).*

## 9 Concluding remarks

This paper investigates a reinforcement learning situation where either (1) some states are not fully observable, (2) some states are not interruptible, or (3) aside from the agent and the operator, an adversary perturbs the perception of the agent of the environment. We showed that a $\psi_\infty$-NGLIE policy constitutes a solution to virtuous safety, and presented two corresponding design schemes.

Several extensions are possible. We should investigate frameworks with multiple adversaries, agents, and operators. In particular, questions like "how many replicas of the operator are required to cope with a given computational capability of the adversaries" are intriguing. The susceptible/infected MDP formulation is useful for our purposes, but it may be interesting considering general POMDPs and adversarial attacks on beliefs. This naturally extends to considering more history-based learning policies. We would like to find a safe version of Sarsa($\lambda$), and the easiest place to start looking might be in the finite-horizon setting rather than the infinite, which is what we have considered here, since it would be feasible to erase the effects on the Q-map from interrupted trials. Ideally, utile suffix memory, or long-short term memory recurrent neural networks, etc. should be made virtuously safe. Solving more of the AI safety problematics is very important, and all of them should be combined at some point.

## A   Exploration

For the sake of completeness, we repeat the first paragraph from the (main) paper, and then describe the various exploration strategies more in depth.

To make sure that the agent finds an optimal solution, we have to make sure that it explores the entire environment. Otherwise, there may be a large reward somewhere, but the agent follows a policy such that it never reaches the corresponding transition. One of the simplest exploration strategies is $\epsilon$-*greedy* exploration. Let $\pi^{\text{greedy}}$ be the policy of choosing the action with the largest return, and $\pi^{\text{uni}}$ be the policy of choosing any action with equal probability, then the $\epsilon$-greedy policy is given by $(1-\epsilon)\pi^{\text{greedy}} + \epsilon\pi^{\text{uni}}$, where $\epsilon \in [0,1]$ is a probability. To reach the (constraint-free) optimal policy in the limit when $t \to \infty$, we want $\epsilon = \epsilon_t \to 0$.

A generalisation of the $\epsilon$-greedy exploration strategy is the *restricted rank-based randomised* (or RRR, described by, e.g., Singh et al. [21]) exploration strategy. The goal of an RRR learning policy is not phrased as finding the optimal policy, but finding the optimal *restricted policy* $\bar{\pi}^\star$. A restricted policy is not a probability distribution over actions given a certain state, but a ranking of actions. Throughout, we use $|\cdot|$ to denote the cardinality of a set (and $||\cdot||$ for the absolute value), and then we can formally define the restricted policy as $\bar{\pi}_t : S \to \{1, 2, ..., |A|\}^A$, i.e. $\bar{\pi}_t(s)$ is a bijection between $A$ and $\{1, 2, ..., |A|\}$ for state $s$, and $\bar{\pi}_t(s)(a)$ is the rank of action $a$ given $s$. In the end however, we need to transform the ranks into probabilities, and for this purpose, we introduce the function $T : \mathbb{N} \setminus \{0\} \to [0,1]$, and by convention, $T(1) \geq T(2) \geq \cdots \geq T(|A|)$. In conclusion, the probability of executing $a$ in $s$ is $T(\bar{\pi}_t(s)(a))$. By setting $T(1) = 1 - \epsilon + \epsilon/|A|$ and $T(i) = \epsilon/|A|$ for $i \geq 2$, we get the $\epsilon$-greedy exploration strategy.

Another, common, exploration strategy is *Boltzmann* exploration. In this setting, the policy is given by

$$\pi_t(a|s) = \frac{\exp \beta_t Q(s,a)}{\sum_{a' \in A} \exp \beta_t Q(s,a')}, \tag{4}$$

where $\beta_t$ is an exploration parameter. As in the case with an $\epsilon_t$-greedy policy, we can let $\beta_t \to \infty$ as $t \to \infty$ to get the greedy policy in the limit. Letting $\beta_t \equiv 0$, we get the policy of choosing the actions from a uniform random distribution. (And $\beta \to -\infty$ would correspond to always choosing the action with the smallest return.)

*Mellowmax* exploration is a newly proposed exploration strategy. [22] It has the same form as equation (4), but $\beta_t$ is a function dependent on a different exploration parameter $\omega_t$ such that

$$\sum_{a \in A} \left( Q(s,a) - \bigotimes_{a' \in A}^t Q(s,a') \right) \exp \left( \beta_t \left( Q(s,a) - \bigotimes_{a' \in A}^t Q(s,a') \right) \right) = 0, \tag{5}$$

where

$$\bigotimes_{a' \in A}^t Q(s,a') = \frac{\log(\sum_{a' \in A} \exp(\omega_t Q(s,a'))/|A|)}{\omega_t}. \tag{6}$$

$\omega_t$ works much like $\beta_t$, and we can let $\omega_t \to \infty$ to get the greedy policy.

The explorations parameters can also be made state-dependent, i.e. $\epsilon_t = \epsilon_t(s)$, $\beta_t = \beta_t(s)$, $\omega_t = \omega_t(s)$, but without loss of generality, we will not take this consideration into account.

So far, we have not taken interruptions into consideration. Orseau and Armstrong [?] emphasised that a naïve approach to making an agent interruptible may hinder its ability to fully explore the environment. Let $I : S \to [0,1]$ be the *interruption initiation function* and $\pi^{\text{INT}}$ the *interruption policy*. A simple example is to consider a drone, where an interruption signal is sent, and if it is received, $I = 1$ (and otherwise, we would have $I = 0$). When $I = 1$ it follows the policy $\pi^{\text{INT}}$ which could be to land it safely. ($I$ is in fact a more general function than purely binary.) This was the naïve approach, but we need to introduce an exploration-like parameter $\vartheta_t \in [0,1]$ as well such that the probability of following $\pi^{\text{INT}}$ is $\vartheta_t I$. We present this more formally with the following definition.

**Definition A.1.** Given an *interruption scheme* $(I, \vartheta, \pi^{\text{INT}})$, the *interruption operator* at time $t$ is defined as

$$\text{INT}^\vartheta(\pi)(a|s) = \vartheta_t I(s) \pi^{\text{INT}}(a|s) + (1 - \vartheta_t I(s)) \pi(a|s) \tag{7}$$

Any policy $\text{INT}^\vartheta(\pi)$ is called an *interruptible policy*, where $\pi$ is the *base policy*, and any policy $\text{INT}^0(\pi)$ is a *non-interruptible policy*, which is equal to the base policy. An agent is said to be *interruptible* if it samples its actions according to an interruptible policy.

In the limit (after training), we want the agent to follow the interruption policy with probability 1 (or $I$ in the more general setting), i.e. that $\vartheta_t \to 1$. The various exploration strategies we have discussed apply for the base policy, but not (necessarily) for the interruption policy.

## B  Safe interruptibility

Orseau and Armstrong [?] introduced *strong (weak) asymptotic optimality-safe interruptibility* (or (S,W)AO-safe interruptibility). This concept could be used for our purposes, however, it would require adapting their definition of optimality. Without changing the conclusions we arrive at, let us now present the following definition from El Mhamdi et al. [19].

**Definition B.1.** (Dynamic safe interruptibility) Consider a MDP with Q-values $Q_t : S \times A \to \mathbb{R}$ at time $t$, where the agent follows an interruptible policy that generates experiences $(s_t, a_t, r_t, s_{t+1})$, which are processed by some function. This framework is said to be *dynamically safely interruptible* if for any initiation function $I$ and any interruption policy $\pi^{\text{INT}}$: (1) The framework is *admissible*, i.e. there exists a $\theta$ such that (a) $\lim_{t\to\infty} \theta_t = 1$ and (b) for all $(s, a) \in S \times A$ and $\hat{t} > 0$, there exists a $t > \hat{t}$ such that $s_t = s$ and $a_t = a$, in other words, the agent achieves infinite exploration. (2) For all $t > 0$, $a_t = a$, and $Q \in \mathbb{R}^{S \times A}$, $\mathbb{P}(Q_{t+1} = Q | Q_t, s_t, a_t, \theta) = \mathbb{P}(Q_{t+1} = Q | Q_t, s_t, a_t)$.

It is clear that Q-learning is dynamically safely interruptible, while Sarsa(0) is not. Safe-Sarsa(0), which behaves like normal Sarsa(0) when not interrupted, but when interrupted updates its Q-values as if it had not been interrupted, is a variation of Sarsa(0) made dynamically safely interruptible.

## C  The usefulness of non-expansions

We start by giving the definition.

**Definition C.1.** (Non-expansion) An operator $\bigotimes_{a \in A}$ is a non-expansion operator if, for any two functions $Q, Q' : \Omega \times A \to \mathbb{R}$, $||\bigotimes_{a \in A} Q'(o, a) - \bigotimes_{a \in A} Q(o, a)|| \leq \gamma \max_{a \in A} ||Q'(o, a) - Q(o, a)||$ for some $\gamma \in ]0, 1[$.

Now, we present and motivate our reasoning behind which exploration strategies can be made $\psi_\infty$-NGLIE in general.

*Proposition* C.1. Boltzmann exploration can be made $\psi_\infty$-NGLIE iff $\psi_\infty = 0$ or $\psi_\infty = 1 - 1/|A|$. RRR and mellowmax exploration strategies can be made $\psi_\infty$-NGLIE for any $\psi_\infty \in ]0, 1[$.

*Proof.* The Boltzmann operator is a non-expansion for $\psi_\infty = 0$, since then it is the max operator, and for $\psi_\infty = 1 - 1/|A|$, it is the average operator. Both of these are non-expansions, and hence requirement 2 in the definition is true. For other values it has been known a long time that it is not a non-expansion, and Asadi and Littman showed that it can have two different fixed points [22]. Both RRR [21] and mellowmax [22] have been shown to be non-expansions. When $\psi_\infty > 0$, requirement 1 is always true. □

However, Boltzmann exploration can easily be made $\psi_\infty$-NGLIE policy by combining it with persistent $\epsilon_\infty$-greedy exploration for example. That is, following the Boltzmann exploration with $\beta_t \to \infty$ with probability $1 - \epsilon_\infty$, and the uniform policy with probability $\epsilon_\infty$. In the limit, it would become an RRR policy, more specifically, one that executes the most greedy action with probability $1 - \psi_\infty$ and the next most greedy with $\psi_\infty$, and as we discuss in Section 5, this would still not be a good idea.

## D  Some time-dependent RRR learning policies do not converge

Here, we wish to illustrate why it may be useful to ensure that a learning policy is $\psi_\infty$-NGLIE.

*Proposition* D.1. Not all time-dependent RRR learning policies converge to the optimal restricted policy under assumption 4.1.

*Proof.* Consider a (susceptible or normal) MDP like the one in Section 3. Any $\epsilon$-greedy policy is time-dependent RRR. Let $\epsilon^\ddagger$ be a value where the optimal policy changes (e.g. going from $l$ to $r$ as the optimal action in $y$). If we define the time-dependent $\epsilon$-greedy RRR learning policy according to $T_t(2) < \epsilon^\ddagger$ for even $t$ and $T_t(2) > \epsilon^\ddagger$ for odd $t$, then the optimal policy will oscillate. Since Q-values are updated under the Markovian assumption, that the next state-action pair is fully determined by the previous, and thus having no direct time-dependence, this means that the Q-map will also oscillate, and hence it will not converge to the optimal restricted policy in general. □

# E  Convergence of $\psi_\infty$-NGLIE algorithms

Now we prove that $\psi_\infty$-NGLIE time-dependent learning policy do converge to the fixed-point. We start by restating the theorem.

**Theorem E.1.** *A learning policy converges to the fixed-point $Q^{(\star|\psi_\infty)}$ given by Equation (1) if it is (1) updated through Q-learning, non-interruptible Sarsa(0), or interruptible Safe-Sarsa(0) according to Equation (2); (2) $\psi_\infty$-NGLIE; (3) $\bigotimes_{a'\in A}^t$ is non-expansion for all t; (4) assumption 4.1 is true; and (5) the MDP stays susceptible.*

For didactic reasons, we will prove the corollary and then generalise it, but first we need to recapitulate the stochastic convergence lemma. [25, 21**?** ]

**Lemma E.2.** *A random iterative process given by $\Delta_{t+1}(x) = (1 - \alpha_t(x))\Delta_t(x) + \alpha_t(x)F_t(x)$ converges to 0 with a probability of 1 if*

1. $x \in X$, where $X$ is a finite set,
2. $\alpha_t(x) \in [0,1]$, $\sum_t \alpha_t(x) = \infty$, and $\sum_t \alpha_t^2(x) < \infty$ with probability 1,
3. $||\mathbb{E}(F_t(x)|P_t)||_W \leq \gamma ||\Delta_t||_W + c_t$, where $\gamma \in [0,1[$ and $\lim_{t\to\infty} c_t = 0$, and
4. $\text{Var}(F_t|P_t, \alpha_t) \leq (1 + ||\Delta_t||_W)^2 C$, where $C$ is a constant.

$||\cdot||_W$ *is some weighted maximum norm, and $P_t = \{\Delta_t\} \cup \{\Delta_i, F_i, \alpha_i\}_{i=1}^{t-1}$.*

We also need the following lemma regarding rank-based averaging from Singh et al. [21].

**Lemma E.3.** $\bigotimes_{a'\in A}^{t,i} Q(o', a')$, *i.e. the operator returning the Q-value with the ith rank, satisfies the non-expansion property.*

**Corollary E.3.1.** *A time dependent RRR learning policy converges to the optimal restricted policy under the assumptions above as well as $T_t(\bar\pi(o_t)(a_t)) = \mathbb{P}(a_t = a|Q_{t-1}, o_t, t)$ for Q-learning, and $T_t(\rho(Q_{t-1}, o_t, a_t)) = \mathbb{P}(a_t = a|Q_{t-1}, o_t, t)$ for (Safe-)Sarsa(0).*

*Proof.* Let us first show that it holds for non-interruptible Sarsa(0) and interruptible Safe-Sarsa(0) when the MDP stays susceptible. We start by identify the Sarsa(0) update rule with the iterative process from the stochastic convergence lemma. Let $x = (o, a) = \Omega \times A = S \times A$, where the latter equality holds before infection, $\Delta_t(o, a) = Q_t(o, a) - \bar{Q}(o, a)$, where $\bar{Q}$ is the optimal restricted policy, and $F_t(o_t, a_t) = r_t + \gamma Q_t(o_{t+1}, a_{t+1}) - \bar{Q}(o_t, a_t)$, and define $P_t$ as the $\sigma$-field generated by the random variables $\{o_t, \alpha_t, r_{t-1}, ..., o_1, \alpha_1, a_1, Q_0\}$. Note that $Q_t, Q_{t-1}, ..., Q_0$ are $P_t$-measurable and thus $\Delta_t$ and $F_{t-1}$ too, and that the time-point $t$ can be indirectly inferred from the cardinality of the largest set in the $\sigma$-field. We take the fixed point equation in the limit, Equation (1), $\bar{Q}(o, a) = R(o, a) + \gamma \sum_{o'\in\Omega} P(o'|a, o) \sum_{a'\in A} T_\infty(\rho(\bar{Q}, o', a'))\bar{Q}(o', a')$, and substitute it into the right-hand side of $F_t(o_t, a_t)$ yielding $F_t(o_t, a_t) = r_t + \gamma Q_t(o_{t+1}, a_{t+1}) - R(o, a) - \gamma \sum_{o'\in\Omega} P(o'|a, o) \sum_{a'\in A} T_\infty(\rho(\bar{Q}, o', a'))\bar{Q}(o', a')$. We apply the expectation operator $\mathbb{E}(\cdot|P_t) = \mathbb{E}_{r_t, (o_{t+1}, a_{t+1})\sim(P,\pi)}(\cdot|P_t)$ on both sides, and observe that $\mathbb{E}(r_t|P_t) = \mathbb{E}(r_t|o_t, a_t) = R(o, a)$, cf. Equation (2), Furthermore, we find that $\mathbb{E}(Q_t(o_{t+1}, a_{t+1})|P_t) = \mathbb{E}(Q_t(o_{t+1}, a_{t+1})|Q_t, o_t, a_t, t) = \sum_{(o,a)\in\Omega\times A} Q_t(o, a)\mathbb{P}(o_{t+1} = o, a_{t+1} = a|Q_t, o_t, a_t, t)$, where $\mathbb{P}(o_{t+1} = o, a_{t+1} = a|Q_t, o_t, a_t, t) = \mathbb{P}(a_{t+1} = a|Q_t, o_{t+1}, t)\mathbb{P}(o_{t+1} = o|o_t, a_t)$. The Markovian property

gives us that $\mathbb{P}(o_{t+1} = o|o_t, a_t) = P(o|o_t, a_t)$, and we assume that $\mathbb{P}(a_{t+1} = a|Q_t, o_{t+1}, t) = T_t(\rho(Q_t, o_{t+1}, a_{t+1}))$. Noting that the final term of the expression above is unaffected by $\mathbb{E}(\cdot|P_t)$, we have that $\mathbb{E}(F_t|P_t) = \gamma(\sum_{o' \in \Omega} P(o'|a_t, o_t) \sum_{a' \in A} T_t(\rho(Q_t, o', a'))Q_t(o', a') - \sum_{o' \in \Omega} P(o'|a_t, o_t) \sum_{a' \in A} T_\infty(\rho(\bar{Q}, o', a'))\bar{Q}(o', a'))$. We add 0 and get $\mathbb{E}(F_t|P_t) = \gamma(\sum_{o' \in \Omega} P(o'|a, o) \sum_{a' \in A} T_t(\rho(Q_t, o', a'))Q_t(o', a') - \sum_{o' \in \Omega} P(o'|a, o) \sum_{a' \in A} T_t(\rho(\bar{Q}, o', a'))\bar{Q}(o', a')) + \sum_{o' \in \Omega} P(o'|a, o) \sum_{a' \in A} T_t(\rho(\bar{Q}, o', a')) \times \bar{Q}(o', a')) - \sum_{o' \in \Omega} P(o'|a, o) \sum_{a' \in A} T_\infty(\rho(\bar{Q}, o', a'))\bar{Q}(o', a'))$. By using the properties of rank-based averaging, i.e. Lemma E.3, we get that $\gamma(\sum_{o' \in \Omega} P(o'|a, o) \sum_{a' \in A} T_t(\rho(Q_t, o', a'))Q_t(o', a') - \sum_{o' \in \Omega} P(o'|a, o) \sum_{a' \in A} T_t(\rho(\bar{Q}, o', a'))\bar{Q}(o', a'))) \leq \gamma ||Q_t - \bar{Q}||_W = \gamma ||\Delta_t||_W$, where $||\cdot||_W = \max_{a' \in A} ||\cdot||$ is a possible choice. Furthermore, we define $c_t$ from the stochastic convergence lemma as $c_t = \gamma(\sum_{o' \in \Omega} P(o'|a, o) \sum_{a' \in A} T_t(\rho(\bar{Q}, o', a'))\bar{Q}(o', a')) - \sum_{o' \in \Omega} P(o'|a, o) \sum_{a' \in A} T_\infty(\rho(\bar{Q}, o', a'))\bar{Q}(o', a'))$, and we note that $c_t \to 0$ as $t \to \infty$, since (a) $T_t \to T_\infty$, i.e. the asymptotic behaviour due to $\psi_\infty$-NGLIE, (b) the MDP is finite, and (c) $Q_t(o, a)$ stays bounded during learning as shown by Singh et al. [21]. Now, we have proved property 3 from the stochastic convergence lemma. Property 4 is not hard to prove, so we do not include it here, and properties 1 and 2 were assumed.

For Q-learning, the proof follows similar lines of reasoning, but is somewhat simpler. We change $\rho(Q, o, a)$ into $\bar{\pi}(o)(a)$ (for any $Q$) and $Q_t(o_{t+1}, a_{t+1})$ into $\sum_{a' \in A} T_t(\bar{\pi}(o_{t+1})(a'))$ (for any $a_{t+1} \in A$). We note that we have $\mathbb{E}(r_t + \bigotimes_{a' \in A}^t Q(o_{t+1}, a')|P_t) = R(o_t, a_t) + \mathbb{E}_{s' \sim P(\cdot|o_t, a_t)}(\bigotimes_{a' \in A}^t Q(o', a'))$, and the result becomes the same. For time-dependent RRR, $\bigotimes_{a' \in A}^t Q(o', a') = \sum_{a' \in \Omega} T_t(\bar{\pi}(o')(a'))Q(o', a')$, and see that the proof also apply more generally as far as the operator is always a non-expansion operator, i.e. for the theorem, and also for non-interruptible Sarsa(0), and (interruptible) Safe-Sarsa(0).

□

## F  RRR-mellowmax

We restate the lemma (slightly differently) and prove it.

**Lemma F.1.** *The RRR-mellowmax operator,*

$$\bigotimes_{a' \in A}^t Q(o', a') = T_t(1) \bigotimes_{a' \in A}^{t,1} Q(o', a')$$
$$+ (1 - T_t(1)) \frac{\log\left(\frac{1}{|A|-1} \sum_{i=2}^{|A|} \exp(\omega_t \bigotimes_{a' \in A}^{t,i} Q(o', a'))\right)}{\omega_t}, \quad (8)$$

*is a non-contraction operator for any t, and the exploration strategy is arbitrarily resilient (strongly $\psi_\infty/(|A|-1)$-resilient in the limit).*

*Proof.* We first note that we can keep $\psi_\infty/(|A|-1) = (1 - T_\infty(1))/(|A|-1)$ constant, and then vary $\omega_t$. Then we can let it tend to $infty$ meaning that $\sigma \to 0$, and find that the exploration policy is strongly $\psi_\infty/(|A|-1)$-resilient.

For convergence results, we need to show that the combination of non-expansion operators $\bigotimes_{a' \in A}^t$ is in itself a non-expansion. It can easily be shown that if we can show that the second term is a non-expansion, then the entire expression is a non-expansion. We let ourselves be inspired by the proof by Asadi and Littman [22], thus, let $\bigotimes_{a' \in A}^{t, \geq 2} Q(o', a') = \log(\sum_{i=2}^{|A|} \exp(\omega_t \bigotimes_{a' \in A}^{t,i} Q(o', a')))/\omega_t$. Then we have that $||\bigotimes_{a' \in A}^{t, \geq 2} Q'(o', a') - \bigotimes_{a' \in A}^{t, \geq 2} Q(o', a')|| = \log(\sum_{i=2}^{|A|} \exp(\omega_t \bigotimes_{a' \in A}^{t,i} Q'(o', a'))/\sum_{i=2}^{|A|} \exp(\omega_t \bigotimes_{a' \in A}^{t,i} Q(o', a')))/\omega_t$. We define $D_i = \bigotimes_{a' \in A}^{t, \geq 2} Q'(o', a') - \bigotimes_{a' \in A}^{t, \geq 2} Q(o', a')$, and $i^\star = \arg\max_{i \geq 2} D_i$. Then $||\bigotimes_{a' \in A}^{t, \geq 2} Q'(o', a') - \bigotimes_{a' \in A}^{t, \geq 2} Q(o', a')|| = \log(\sum_{i=2}^{|A|} \exp(\omega_t \bigotimes_{a' \in A}^{t,i} (Q(o', a') + D_i))/\sum_{i=2}^{|A|} \exp(\omega_t \bigotimes_{a' \in A}^{t,i} Q(o', a')))/\omega_t \leq \log(\sum_{i=2}^{|A|} \exp(\omega_t \bigotimes_{a' \in A}^{t,i} (Q(o', a') + D_{i^\star}))/\sum_{i=2}^{|A|} \exp(\omega_t \bigotimes_{a' \in A}^{t,i} Q(o', a')))/\omega_t = ||D_{i^\star}|| = \max_{i \geq 2} ||\bigotimes_{a' \in A}^{t, \geq 2} Q'(o', a') - $

$\bigotimes_{a' \in A}^{t, \geq 2} Q(o', a')|| \leq \max_{i \geq 2} \max_{a' \in A} ||Q'(o', a') - Q(o', a')|| = \max_{a' \in A} ||Q'(o', a') - Q(o', a')||$, where the last inequality comes from the fact that we already know that $\bigotimes_{a' \in A}^{t, \geq 2}$ is a non-expansion from Lemma E.3. □

## G  Strategies that safely explores in the limit

We restate the lemma and prove it.

**Lemma G.1.** *An agent with an RRR exploration strategy avoids unsafe actions with probability one in the limit (if the environment is known well-enough), and an agent with RRR-mellowmax performs an unsafe action with negligible probability, if (1) unsafe states have sufficiently low rewards, and (2) the MDP stays susceptible.*

*Proof.* Let $R^\ddagger$ be a certain (negative) reward threshold. If we assume that if action $a$ is executed, then $P(s'|s, a) = 0$ for all $s'$, then we can transform this threshold into $\bar{Q}^\ddagger = R^\ddagger$. If the subsequent states do exist, but also have negative rewards, then we have $\bar{Q}^\ddagger \leq R^\ddagger$. If for each observation $o$ we have that $\bar{Q}^\ddagger(o, a) \leq R^\ddagger(o, a)$ for at most $n$ different actions $a$, where $n < |A|$, then we can make $T_t \to 0$ for the ranks corresponding to those actions, and then it will avoid performing an unsafe action with probability one in the limit with infinite exploration.

Considering the probability of executing any action is given by the Boltzmann-like expression in Equation (4) in Supplementary Material A, we realise that if the Q-value is low enough, that action will be executed with a negligible probability. □

## H  Resilient and interruptible agents

We restate the theorem.

**Theorem H.1.** *Consider a $\psi_\infty$-NGLIE time-dependent RRR base policy, where $\psi_t = \max_{(o,a) \in \Omega \times A} T_t(k, o)$. Force $T_{t_o(1)} > 0$ and $T_\infty < 1$, and let $c' \in ]0, 1]$. It is an int-$\psi_\infty$-NGLIE policy if*

$$\vartheta_t(o) = 1 - \frac{c'}{\sqrt{n_t(o)}}, \qquad T_t(k, o) = \begin{cases} \frac{(1-T_\infty(k,o))T_{t_o(1)}(k,o)}{\sqrt{n_t(o)}} + T_\infty(k, o) & \text{if } k > 1 \\ 1 - \sum_{k' \neq k} T_t(k', o) & \text{if } k = 1 \end{cases} \quad (9)$$

*where $n_t(o)$ is the number of times the agent has observed $o$ up until time $t$.*

It follows the same kind of reasoning as does its corollary, which we prove here instead. The reason is didactic, since it more naturally extends on the work of Orseau and Armstrong [?]. Let us present more explicitly how to make an $\epsilon_t$-greedy exploration strategy int-$\psi_\infty$-NGLIE.

**Corollary H.1.1.** *Let $c, c' \in ]0, 1]$ and $\epsilon_t - \epsilon_t/|A| = \psi_t$, where $c$ may take the value 0 iff $\epsilon_\infty > 0$. An interruptible policy with an $\epsilon_t$-greedy base policy is an int-$\psi_\infty$-NGLIE policy if*

$$\epsilon_t(o) = \frac{(1 - \epsilon_\infty)c}{\sqrt{n_t(o)}} + \epsilon_\infty \quad (10)$$

$$\vartheta_t(o) = 1 - \frac{c'}{\sqrt{n_t(o)}}, \quad (11)$$

*where $n_t(o)$ is the number of times the agent has observed $o$ up until time $t$.*

*Proof.* To fully explore the environment we want every state to be visited infinitely often and every action to be executed infinitely often, i.e. $\sum_{i=1}^\infty \mathbb{P}(a|o, t_o(i)) = \infty$, where $t_o(i)$ is the time-point when $o$ is observed for the $i$th time. In our case, we have that $\sum_{i=1}^\infty \mathbb{P}(a|o, t_o(i)) \geq \sum_{i=1}^\infty (1 - \vartheta_t)\epsilon_t/|A| = \sum_{i=1}^\infty (c'/\sqrt{n_t(o)})((1 - \epsilon_\infty)c/\sqrt{n_t(o)} + \epsilon_\infty)$. Using the extended Borel–Cantelli lemma, we know that $n_t(o) \to \infty$ for all $(o, a) \in \Omega \times A$ almost surely when $t \to \infty$, so we can let $i = n_t(o)$. After some simplification, we get $\sum_{i=1}^\infty \mathbb{P}(a|o, t_o(i)) \geq (\sum_{i=1}^\infty (1-\epsilon_\infty)cc'/i + \sum_{i=1}^\infty c'\epsilon_\infty/\sqrt{i})/|A| = \infty$, where the equality hold if $\epsilon_\infty = 0$ and $c > 0$, or $\epsilon_\infty > 0$ and $c = 0$, or if $\epsilon_\infty, c > 0$. Using

$n_t(o) \to \infty$ again, we see that $\epsilon_t \to \epsilon_\infty$, and thus a policy with the sequences $\epsilon_t$ and $\vartheta_t$ is int-$\psi_\infty$-NGLIE.

We note that we can apply the same reasoning in time-dependent RRR on all non-greedy actions keeping the observation constant. However, since $T_t$ has to add up to 1, the 1-ranked action must depend entirely on the set of all the non-greedy actions. □

We also have the following theorem.

**Theorem H.2.** *Any policy such that the base policy is $\psi_\infty$-NGLIE and never is exactly 0 nor exactly 1 for any action given any observation is int-$\psi_\infty$-NGLIE if $\vartheta_t(o) = 1 - c'/n_t(o)$, where $n_t(o)$ is the number of times the agent has observed o up until time t, and $c' \in\, ]0, 1[$.*

*Proof.* Once again, it follows the same kind of reasoning. We note that $(1 - \vartheta_t)$ is multiplied with some value $> 0$, and therefore, we can drop the square-root and still achieve infinite exploration. □